\documentclass[11pt,a4paper]{article}
\usepackage[breaklinks=true]{hyperref}
\usepackage[hyperref]{acl2021}
\usepackage{times}
\usepackage{latexsym}
\usepackage{bm}
\usepackage{multirow}

\usepackage{tikz,pgfplots}%
\usepackage{amsmath,amssymb}
\usepackage[colorinlistoftodos]{todonotes}
\usetikzlibrary{shapes.geometric, arrows}
\usepackage{float}

\pgfplotsset{colormap={whiteblack}{color=(white) color=(black)}}

\usepackage{microtype}

\aclfinalcopy 

\title{Continuous representations of intents for dialogue systems}

\author{Sindre Andr\'{e} Jacobsen \\
  Department of Computer Science\\
  University of Sheffield\\
  211 Portobello,
  Sheffield S1 4DP, UK\\
  \texttt{sinder33@gmail.com} \\\And
  Anton Ragni \\
  Department of Computer Science\\
  University of Sheffield\\
  211 Portobello, Sheffield S1 4DP, UK\\
  \texttt{a.ragni@sheffield.ac.uk}\\}

\date{}

\begin{document}
\maketitle
\begin{abstract}
Intent modelling has become an important part of modern dialogue systems. With the rapid expansion of practical dialogue systems and virtual assistants, such as Amazon Alexa, Apple Siri, and Google Assistant, the interest has only increased. However, up until recently the focus has been on detecting a fixed, discrete, number of seen intents. Recent years have seen some work done on unseen intent detection in the context of zero-shot learning. This paper continues the prior work by proposing a novel model where intents are continuous points placed in a specialist Intent Space that yields several advantages. First, the continuous representation enables to investigate relationships between the seen intents. Second, it allows any unseen intent to be reliably represented given limited quantities of data. Finally, this paper will show how the proposed model can be augmented with unseen intents without retraining any of the seen ones. Experiments show that the model can reliably add unseen intents with a high accuracy while retaining a high performance on the seen intents.%
\end{abstract}

\section{Introduction}
Dialogue systems and virtual assistants have started becoming successful in recent years. While the technology is evolving at great speeds it still is a long way before they can handle a truly natural conversation \citep{turning-test-book}. These systems are growing in features, possible ways to interact with the user and are increasingly multilingual. This process from the user communicating with the machine to getting a response can be broken down into multiple steps. This paper is looking into one of these steps called intent prediction that aims to determine the user's goal.%

A common approach to intent prediction consists of using a machine learning model capable of mapping user queries to a discrete, fixed, intent class. With the recent development in deep learning the interest towards intent prediction in recent years have increased yet the common approach has not changed \citep{user-query-wikipedia, convolution-crf-intent-detection, zhang-etal-2019-joint}. This common approach has serious limitations. One major limitation is that any intent is assumed to belong to a fixed number of intent classes known well in advance. Machine learning models that follow such approach fail to take into account that the number of intents is dynamic (e.g. a rapid emergence of new intents in the wake of COVID-19 pandemic). These models would have to be retrained or significantly adapted to add new intents. Furthermore, they are incapable of finding relationships between existing and emerging intents which might be important for both the users and the developers.%

There have been an increased interest in working with models that can handle unseen intents to tackle some of the above issues, such as zero-shot learning \citep{zero-shot-attribute-based-classification, synthesied-classifier-zsl}. In this paper we propose a model that can tackle all of those issues. The main idea is to represent intents as points in a continuous space called intent space, which among other benefits allows to compute distances between intents and explore the relationships among seen and unseen intents. This paper shows how these intent spaces can be embedded into standard forms of neural networks adapted to handle a variable number of classes and how unseen intents can be added without retraining any part of the model learnt on seen intents.%

\section{Background}
\label{sec:background}
Intent prediction is a machine learning task where the goal is to predict correct intent class $c$ for a given sentence $\bm X$ consisting of $T$ words. One way to accomplish this is by using a neural network such as Recurrent Neural Network (RNN) \citep{rnn}. RNNs take sentences encoded using vector representation schemes, such as one-hot encoding or word embeddings \citep{word-representation-vector-mikolov}, and recursively compute history states ${\bm h}_t$ for each word ${\bm x}_t$ using the history state ${\bm h}_{t-1}$ of the previous word ${\bm x}_{t-1}$ 
\begin{equation}
{\bm h}_t = {\bm \sigma}({\bm U}{\bm h}_{t - 1} + {\bm V} {\bm x}_t + {\bm b})
\label{eq_rnn}
\end{equation}
where ${\bm U}$, ${\bm V}$, ${\bm b}$ are RNN parameters and ${\bm\sigma}$ is an element-wise non-linearity, such as sigmoid, $\tanh$ or ReLU. More complex forms of RNNs such as LSTM \citep{lstm} and GRU \citep{gru} can also be used.%

We can define the probability of the sentence to belong to intent class $c$ using softmax%
\begin{equation}
P(y=c|{\bm x}_{1:T}) = \frac{\exp\left({\bm a}_{c}^{\sf T} {\bm h}_{T} + d_{c}\right)}{\sum_{c=1}^{C} \exp\left({\bm a}_{c}^{\sf T} {\bm h}_{T}+d_{c}\right)}
\label{eq:softmax}
\end{equation}
where ${\bm h}_{T}$ is the final history vector associated with the final word ${\bm x}_{T}$ and ${\bm A}$, ${\bm d}$ are additional RNN parameters. Note that the equation above generally implies that the number of classes $C$ is known in advance. The RNN parameters ${\bm\theta}=({\bm U}, {\bm V}, {\bm b}, {\bm A}, {\bm d})$ can be optimised on a training set ${\mathcal D}$ to maximise the probability of assigning sentences ${\bm X}_{r}$ to reference intent classes $y_r$%
\begin{equation}
{\mathcal F}({\bm\theta};{\mathcal D}) = -\frac{1}{R} \sum_{r=1}^{R} \log(P(y_{r}|{\bm X}_{r}))
\label{eq_rnn_loss}
\end{equation}
As previously mentioned, this approach to intent modelling comes with several serious limitations.%

\section{Intent Space}
\begin{figure}[H]
\centering
\includegraphics[width=4cm, height=4cm]{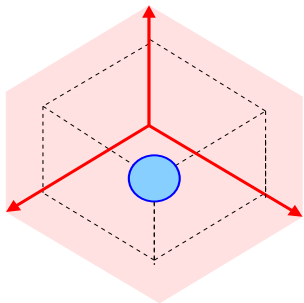}
\caption{Example of three-dimensional intent space}
\label{fig:intent_space}
\end{figure}

One approach to solve these limitations is to introduce an intent space. Intent space is the concept of representing each intent $c$ as a point in a $B$-dimensional space, rather than a discrete class. Figure~\ref{fig:intent_space} shows a simple illustration of a 3-dimensional intent space with one intent present.%
The intents will be assumed to live in a low dimensional space with $B$ bases. The goal is for similar intents to be grouped together, while different ones to have a distance between them. As the intent space is not fixed to a particular number of intents, it will also allow for adding unseen intents and being able to measure the similarity to existing ones. Intent space opens up a range of other possibilities of what can be achieved. The rest of this section will discuss some of them.%

\subsection{Spaces}
There are at least two options for modelling bases (arrows in Fig.~\ref{fig:intent_space}) in this new coordinate system. One such approach would be to adopt a vector space, where bases are vectors ${\bm w}_{1},\ldots,{\bm w}_{B}$ and points (circle in Fig.~\ref{fig:intent_space}) can be expressed using coordinates ${\bm\alpha}_{c}=[\alpha_{c,1}\ldots \alpha_{c,B}]^\top$ as
\begin{equation}
{\bm u}_{c} = \sum_{b=1}^{B} \alpha_{c,b} {\bm w}_{b}
\label{eq_vector_space}
\end{equation}
Vector spaces allow for a simple representation of intents but offer few model parameters to be powerful. To increase modelling power we can model bases using matrices%
\begin{equation}\label{eq_matrix_space}
{\bm U}_{c} = \sum_{b=1}^B \alpha_{c,b} {\bm W}_{b}
\end{equation}
It is also possible to use more complex schemes that change the nature of bases and/or how points are expressed. If a more parameter efficient model is required compared to equation~\eqref{eq_matrix_space}, it is possible to examine reduced rank matrix spaces%
\begin{equation}\label{eq_rank_matrix_space}
{\bm U}_{c} = \sum_{b=1}^{B} \alpha_{c,b} \sum_{k=1}^{K} {\bm w}_{b,k} {\bm w}_{b,k}^{\sf T} 
\end{equation}
This allows to retain a matrix representation albeit of reduced rank $K$ (fewer parameters).%

There are a number of important considerations to be made about bases. One consideration is what these bases should represent. If each intent can be represented as a combination of a small number of {\em eigen} or proto-intents then each basis should represent one of those {\em eigen}-intents. Another consideration is how many bases would be needed for any given set of intents. If intents are unrelated the optimal number of bases $B$ would likely be equal to the number of (seen) intents $C$. An unseen intent that is related to one of the seen intents can then be represented by means of the corresponding basis and possibly other bases as well. An unrelated unseen intent will be forced to either pick the closest basis or make use of all or some available bases. However, if the unseen intent is related to more than one seen intent it should benefit from the rich representation available to it.%

\subsection{Space Expansion}
Different choices of basis representations discussed in the previous section do not alter the low-dimensional nature of the intent space. Such simplified approach may struggle to deal with complex intents that must co-exist within the same space. Thus, it is desirable to be able to expand the space to meet different levels of complexity. One approach to accomplish this is shown below%
\begin{equation}\label{eq_omega}
{\bm U}_{c} = \sum_{b=1}^B \alpha_{c,b} {\bm\Omega}_{c,b} {\bm W}_b
\end{equation}
where ${\bm\Omega}_{c,b}$ is an intent $c$ expansion matrix for basis $b$. Such intent-dependent expansion enables different intents to use either equation~\eqref{eq_omega} or \eqref{eq_matrix_space} depending on the complexity needed for their specific cases. It is also possible to incrementally increase complexity by initially using equation~\eqref{eq_matrix_space} and then continuing with equation~\eqref{eq_omega} where all ${\bm\Omega}_{c,b}$ are initialised as identity matrices.%

\subsection{Coordinates}
Coordinates ${\bm \alpha}=[{\bm\alpha}_{1}\;\ldots\;{\bm\alpha}_{C}]$ are an important feature of the intent space and can be set in a variety of ways. One such way is to initialise them randomly. Interpretability can be improved by representing each ${\bm \alpha}_{c}$ as the one-hot encoding of that intent making ${\bm\alpha}$ an identity matrix. Such a choice implies that the number of bases $B$ is set to the number of seen intents $C$%
\begin{equation}
{\bm\alpha}_{c} = \begin{bmatrix}\alpha_{c,1} & \alpha_{c,2} & \ldots & \alpha_{c,B}\end{bmatrix}^{\sf T}
\end{equation}
If ${\bm\alpha}$ are not constrained in any way the intent-space is {\em Euclidean}. One issue with Euclidean spaces is an overall lack of coordinate interpretability (e.g. negative values, scale) as the interaction between coordinates $\alpha_{c,b}$ and bases ${\bm W}_{b}$ makes interpreting coordinates complicated. However, if each ${\bm\alpha}_{c}$ is forced to lie on a {\em simplex space}%
\begin{equation}
\alpha_{c,b} = \frac{\exp(\beta_{c,b})}{\sum_{b=1}^B \exp(\beta_{c,b})}
\end{equation}
allowing to see how much each coordinate $\alpha_{c,b}$ takes from another intent/basis $b$, and how much each unseen intent learns from the seen ones. Note that ${\bm\beta}$ can be viewed as unnormalised coordinates.%

\subsection{Space Embeddings}
There are multiple ways how intent spaces can be incorporated within modern forms of neural networks. Consider, for example, the RNN which has three model parameters $\bm U$, $\bm V$ and $\bm b$ linked with input sequence embedding, and two model parameters ${\bm A}$ and ${\bm d}$ linked with intent class prediction. Each model parameter serves as the potential candidate for intent space embedding. In this work we will examine model parameters linked with input sequence embedding, which yields three possible options. Regardless of the chosen option, the history state of intent space RNN will become intent-dependent ${\bm h}_{c,t}$ rather than intent-independent ${\bm h}_{t}$. The implications resulting from this change are discussed further in this section.%

The first option is to embed intent spaces into bias vectors $\bm b$ using the vector space approach defined in equation \eqref{eq_vector_space}. This would however only act as an intent-dependent offset with a possibility of anchoring history states associated with different intents in different regions of the history space to aid in intent separability. While this approach could theoretically lead to improved performance, the largest benefit is expected from embedding intent spaces either into the recurrent matrix $\bm U$ or (word) embedding matrix $\bm V$ using the matrix spaces approaches in equations~\eqref{eq_matrix_space}, \eqref{eq_rank_matrix_space} or \eqref{eq_omega}.%

The second option is to introduce intent space into the embedding matrices $\bm V$, which opens up a number of interesting options. One option is to learn discriminative vocabularies. For instance, an intent linked with acquiring information about weather conditions would be expected to learn high-quality embeddings for some words, such as weather and forecast, but not for others, such as Adele and Madonna.%

Finally, it is also possible to introduce intent spaces into model parameters $\bm U$ that control which information from the past is propagated along the sequence. Such intent spaces are expected to be very powerful in controlling which information will be ultimately used for making accurate intent predictions. The update rule used to compute history vector at time $t$ in these intent spaces is given by%
\begin{equation}
{\color{blue}{\bm h}_{c,t}} = {\bm \sigma}({\color{blue}{\bm U}_{c}{\bm h}_{c,t - 1}} + {\bm V}{\bm x}_t + {\bm b})
\end{equation}
where the model parameters ${\bm U}_c$ and the history states ${\bm h}_{t,c}$ are intent-dependent.%

The intent-dependent nature of history states changes how we compute probabilities of sentences to belong to a particular intent class. There are two main options. One approach is to introduce one set of intent-independent parameters. Another approach is to introduce $C$ sets of intent-dependent parameters. Given $C$ history states ${\bm h}_{1,T}$, $\ldots$, ${\bm h}_{C,T}$ associated with each intent, it is possible to compute $C$ scores that reflect how well the sentence ${\bm x}_{1:T}$ matches each intent using just one set of intent-independent parameters ${\bm a}$ and $d$%
\begin{equation}
S({\bm h}_{c,T};{\bm\theta}) = \sigma(\bm{a}^\top \bm{h}_{c,T} + d) \label{eq:probability_intent_space}
\end{equation}
where $\sigma$ is a suitable non-linearity. An alternative approach is to introduce parameters ${\bm a}_{c}$ and $d_c$ for each intent class
\begin{equation}
S({\bm h}_{c,T};{\bm\theta}) = \sigma(\bm{a}_{c}^\top \bm{h}_{c,T} + d_{c}) \label{eq:alternative_probability_intent_space}
\end{equation}
To compute probabilities in either case, we can take the set of (positive) scores and normalise
\begin{equation}
P(y=c|{\bm x}_{1:T}) = \frac{S({\bm h}_{c,T};{\bm\theta})}{\sum_{c} S({\bm h}_{c,T};{\bm\theta})}
\label{eq:intent_space_softmax}
\end{equation}
These two approaches require different numbers of model parameters to be introduced. The former may prove useful in limited resource conditions.%

When a new, previously unseen, intent emerges, it is common to retrain the standard RNN model from scratch. However, the softmax normalisation in equation~\eqref{eq:intent_space_softmax} shared with RNNs is not inherently limited to a fixed number of intent classes. Provided scores associated with the unseen intent do not affect rank ordering and alter relative contributions from other intents on seen sentences, it is possible to retain all existing model parameters. The following section will discuss one practical approach how this can be accomplished, which opens an opportunity for a dynamic model that can add previously unseen intents without the need for altering an already deployed model.%

\subsection{Unseen Intents}
\label{ssec:unseen_intents}
It is hard to envision and cater for all possible user intents in practice. Many intents have a seasonal nature (e.g. flu), some are unexpected (e.g. COVID-19). Thus, handling unseen intents, which includes both unseen intent detection and modelling, is an important practical consideration. Common approaches to unseen intent detection can be split into automatic, semi-automatic and manual. One common automatic approach to decide if a given sentence is likely to belong to an unseen intent class is by using entropy%
\begin{eqnarray}
\lefteqn{{\mathcal H}({\bm X};{\bm\theta}) =}\\
 & & -\sum_{k} P(y=k|{\bm h}_{k,T}) \log\left(P(y=k|{\bm h}_{k,T})\right)\nonumber
\label{eq:entropy}
\end{eqnarray}
The assumption is that the more uncertain the model is about intent classification, the more likely is that the correct intent is in fact unseen. In an idealised setting there exists a threshold $\rho$ such that 
\begin{eqnarray}
{\mathcal H}({\bm X}_{r};{\bm\theta}) &>& \rho \quad \text{if}\;\; y_{r}\;\; \text{is unseen}\\
{\mathcal H}({\bm X}_{r};{\bm\theta}) &\leq& \rho \quad \text{if}\;\; y_{r}\;\; \text{is seen}
\end{eqnarray}
holds true for all sentences ${\bm X}_{r}$ of a dataset ${\mathcal D}$. This might also be a good starting point for developing a semi-automatic approaches that leverage human expertise to make decisions regarding a few pre-filtered candidates. It is also possible to devise intent space based approaches. One such approach could be to obtain an estimate of coordinates $\bm \alpha$ given a sentence ${\bm X}$ and compare that to coordinates of known intents ${\bm\alpha}_{1}$, $\ldots$, ${\bm\alpha}_{C}$. If the estimate ${\bm\alpha}$ is close to one of the known intents ${\bm\alpha}_{c}$, it is  likely to be a seen intent and unseen otherwise.%

When any new intent is identified it needs to be incorporated into the model. It would be beneficial if such a change did not involve altering any of the existing model parameters. Furthermore, we would like to do it in a manner that keeps the accuracy on seen intents high, while also being able to reliably represent any unseen intents. One possible solution to this problem consists of introducing the following regularisation function%
\begin{equation} \label{eq:regularisation_seen_score}
{\mathcal R}({\bm\theta};{\mathcal D}) \!=\! -\frac{1}{KU}\!\! \sum_{r=1}^{K}\! \sum_{u=1}^{U} \log\!\!\left(\!\frac{\max\limits_{c\in[1,C]} S({\bm h}_{r,c,T_{r}};{\bm\theta})}{S({\bm h}_{r,C+u,T_{r}};{\bm\theta})}\!\right)
\end{equation}
which is computed over $K\lll R$ training sentences available to seen intents. The above regularisation function takes into consideration the ratio between seen and unseen intents on the seen training data. This allows us to make sure that none of the unseen intents $C+1, \ldots, C+U$ will significantly impact the performance of classifying sentences into seen intent classes. This is achieved by appropriately weighing the contribution of regularisation function in the overall objective function%
\begin{equation}
{\mathcal O}({\bm\theta};{\mathcal D},{\mathcal D}') = {\mathcal F}({\bm\theta};{\mathcal D}') + \epsilon {\mathcal R}({\bm\theta};{\mathcal D}) + \zeta {\mathcal R}_{{\bm\alpha}}({\bm\theta})
\end{equation}
where ${\mathcal D}$ and ${\mathcal D}'$ are the training sets for seen and unseen intents, ${\mathcal R}_{{\bm\alpha}}({\bm\theta})$ is another regularisation function that either drives the coordinates to the uniform distribution (Simplex spaces) or minimises $L_2$ (Euclidean spaces), $\epsilon$ and $\zeta$ control their respective contributions to the overall objective.%

\subsection{Training}
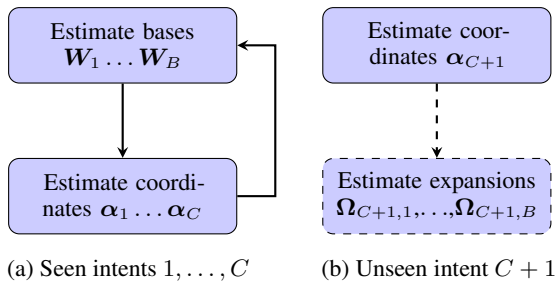
\begin{figure}[H]
\centering
\tikzstyle{block} = [rectangle, rounded corners, minimum width=3cm, minimum height=1cm,text centered, draw=black, fill=blue!20]
\tikzstyle{arrow} = [thick,->,>=stealth]
\begin{minipage}{0.46\columnwidth}
\small
\begin{tikzpicture}[node distance=2cm]
\node (train_w) [block] [text width=2.7cm]{Estimate bases ${\bm W}_{1}\ldots{\bm W}_{B}$};
\node (train_alpha_seen) [block, below of=train_w] [text width=2.7cm]{Estimate coordinates ${\bm\alpha}_{1}\ldots{\bm\alpha}_{C}$};
\draw [arrow] (train_w) -- (train_alpha_seen);
\draw [arrow] (train_alpha_seen.east) -| ++(0.5cm,2.0cm) |- (train_w.east);
\end{tikzpicture}\vspace{0.2cm}\\
(a) Seen intents $1,\ldots, C$
\end{minipage}
\hspace{0.4cm}
\begin{minipage}{0.46\columnwidth}
\small
\begin{tikzpicture}[node distance=2cm]
\node (train_alpha_unseen) [block] [text width=2.7cm]{Estimate coordinates ${\bm\alpha}_{C+1}$};
\node (train_omega) [block, below of=train_alpha_unseen,dashed] [text width=2.7cm]{Estimate expansions ${\bm\Omega}_{C+1,1}$,$\ldots$,${\bm\Omega}_{C+1,B}$};
\draw [arrow,dashed] (train_alpha_unseen) -- (train_omega);
\end{tikzpicture}\vspace{0.2cm}\\
(b) Unseen intent $C+1$
\end{minipage}
\caption{Training methodology for intent space}
\label{fig:intent_space_training}
\end{figure}

The diagram below summarises the overall training process that includes training on seen intents (Fig.~\ref{fig:intent_space_training}~a) followed by an optional training on unseen intents (Fig.~\ref{fig:intent_space_training}~b).%
\vspace{-0.3cm}
As illustrated in the figure above, the intent space parameters ${\bm W}_{1},\ldots,{\bm W}_{B}$ (bases) and ${\bm\alpha}_{1},\ldots,{\bm\alpha}_{C}$ (coordinates) are estimated in an interleaved fashion by first training the bases whilst keeping the coordinates frozen, and then training the coordinates whilst keeping the bases frozen. This process can be repeated for a fixed number of steps, until a certain milestone has been reached (e.g. accuracy, loss), some other criteria have been satisfied (e.g. no significant improvement in loss/accuracy) or a combination of these. Note that the training in figure (Fig.~\ref{fig:intent_space_training}~a) is completed once, while   the training in figure (Fig.~\ref{fig:intent_space_training}~b) can be completed multiple times without any retraining of the model parameters introduced previously.%

To extend the intent-space with one or more previously unseen intents, only the new sets of coordinates need to be estimated. Given a sufficiently large regularisation constant $\epsilon$, the regularisation function in equation~\eqref{eq:regularisation_seen_score} will ensure that the new coordinates will not alter the top-1 rank ordering of the unnormalised scores on a sample of seen intent sentences. If we are satisfied with the results of what we can achieve by introducing only coordinates the process of expanding the intent space with the unseen intents stops. However, if we are not satisfied we can expand the intent space by means of basis-dependent matrices in equation~\eqref{eq_omega}.%

\section{Related work}
The intent space proposed in this work is closely related to Zero-Shot Learning (ZSL), which aims to detect and/or classify examples of classes that either have not been present during training or have no labelled data associated with them. The ZSL has gained a lot of traction within the image domain  \citep{zero-shot-cross-modal-transfer, synthesied-classifier-zsl, devise}. In recent years there have been interest within the natural language processing domain and intent modelling in particular. One simple ZSL approach examined for intent modelling consists of creating or learning representations for given sentence and each intent, and utilising suitable distances to perform classification \citep{zero-shot-lstm, zero-shot-heterogeneous-overlapping-domains}. Both information retrieval style approaches and recurrent neural networks have been examined for learning representations. A similar idea have been explored within the image domain \citep{zero-shot-attribute-based-classification}. Recently there has been interest in using capsule networks for extracting powerful sentence and intent representations \citep{zero-shot-disentangled-intent-representation, zero-shot-capsule, zero-shot-capsule-reconstructed}. One common theme in these papers is that they all require availability of intent labels unlike the proposed intent space. The overall idea of generating a continuous space representation has been examined for representing speakers \citep{cluster-adaptive-training-of-hidden-marklov-models} and languages \citep{statistical-parametric-speech-synthesis, language-space-representation} in speech tasks. Unlike those representations embedded into simpler hidden Markov models, the intent space is embedded into neural networks, which offers access to all the recent developments in deep learning.%

\section{Experiments}
\subsection{Experimental setup}
We conducted experiments on two datasets SNIPS \citep{snips} (primary) and Airline Travel Information System (ATIS) \citep{atis-1990} (secondary). SNIPS is a balanced dataset \footnote{https://github.com/sonos/nlu-benchmark} consisting of 7 intents that represent different domains. There are approximately 2000 sentences available for each intent. We moved the first 100 sentences per intent to create a validation dataset. In experiments where we explored unseen intents one intent was excluded from the training set. ATIS  is an imbalanced dataset within the flight domain made available through Microsoft Cognitive Toolkit \footnote{https://www.kaggle.com/siddhadev/ms-cntk-atis}, where one of the intents, {\tt flight}, comprises approximately seventy percent of the data. Any intent that did not appear in both training and test set was removed to yield a total of 16 intents.%

\textbf{Implementation details}
We use Glove \citep{glove}, which was pre-trained on large quantities of data and features 1.9 million word vocabulary \footnote{ https://nlp.stanford.edu/projects/glove/}, to provide word embeddings with a dimensonality of 300. Dimensionality of all history states was also 300. Words without an embedding were mapped to the average word embedding. The embedding matrix $\bm V$ was initialised to an identity matrix to use unmodified word embeddings on the very first iteration. The coordinates ${\bm\alpha}$ were initialised as one-hot encodings (one basis per intent). The remaining parameters were initialised randomly. All models were trained on a CPU. No grid search or other elaborate methods were used for careful tuning of hyperparameters. such as the learning rate or the modifier term in equation \eqref{eq:regularisation_seen_score}. It is expected that better results may be obtained with more careful tuning.%

RNN was implemented using equation \eqref{eq_rnn} and optimising equation \eqref{eq_rnn_loss} to learn parameters. Intent space was trained using equation \eqref{eq_matrix_space} by following the process illustrated by Figure \ref{fig:intent_space_training}, where we trained bases parameters ${\bm W}$ for 5 epochs, then $\bm\alpha$ for 5 epochs and so on. 
Both approaches made use of the validation set to implement early stopping. When training coordinates or expansion matrices ${\bm\Omega}$ for unseen intents $\epsilon$ was set to 0.20 to penalise the ratio between seen/unseen intents on training data and $\zeta$ was set to 1.00. Stochastic gradient descent (SGD) method with a weight decay was used in SNIPS, while in ATIS Adam optimiser \citep{adam-optimiser} was used to speed up the process. Experiments were forced to stop after epoch $50$, $150$ or $500$ when training seen intents, coordinates or expansion matrices respectively on rare occasions that the early stopping was not triggered.%

\subsection{Intent space}
\begin{figure}[H]
\centering
\begin{tikzpicture}
\begin{axis}[xlabel=Step, ylabel=Accuracy, height=5cm,width=7.5cm, legend pos=south east, legend style={nodes={scale=0.7}},xmin=0, xmax=33,xtick={0, 5, 10, 20, 30}, legend style={at={(1.00,0.06)},anchor=south east}]
\addplot[mark=*, color=blue, mark size=2pt,forget plot] table [x=Step,y=Accuracy,col sep=comma] {data/experiment1_1.csv};
\addplot[mark=square*, color=blue, mark size=2pt,forget plot] table [x=Step,y=Accuracy,col sep=comma] {data/experiment1_2.csv};
\addplot[mark=*, color=blue, mark size=2pt,forget plot] table [x=Step,y=Accuracy,col sep=comma] {data/experiment1_3.csv};
\addplot[mark=diamond*, color=cyan, mark size=3.3pt,forget plot] table [x=Step,y=Seen_Accuracy,col sep=comma] {data/experiment3_1.csv};
\addplot[mark=*, color=cyan, mark size=2.5pt,forget plot] table [x=Step,y=Seen_Accuracy,col sep=comma] {data/experiment2_1.csv};
\addplot[mark=*, color=pink, mark size=2.5pt,forget plot] table [x=Step,y=Unseen_Accuracy,col sep=comma] {data/experiment2_1.csv};
\addplot[mark=diamond*, color=pink, mark size=3.3pt,forget plot] table [x=Step,y=Unseen_Accuracy,col sep=comma, pink] {data/experiment3_1.csv};
\addplot[color=black, mark=*, mark size=2pt] table [x=Step,y=Seen_Accuracy,col sep=comma] {data/dummy.csv};
\addlegendentry{${\bm\alpha}$};
\addplot[color=black, mark=square*, mark size=2pt] table [x=Step,y=Seen_Accuracy,col sep=comma] {data/dummy.csv};
\addlegendentry{${\bm W}$}
\addplot[color=black, mark=diamond*, mark size=3pt] table [x=Step,y=Seen_Accuracy,col sep=comma] {data/dummy.csv};
\addlegendentry{${\bm\Omega}$}
\addplot[color=blue, mark=, mark size=3pt] table [x=Step,y=Seen_Accuracy,col sep=comma] {data/dummy.csv};
\addlegendentry{Seen (top)}
\addplot[color=red, mark=, mark size=3pt] table [x=Step,y=Seen_Accuracy,col sep=comma] {data/dummy.csv};
\addlegendentry{Unseen (bottom)}
\end{axis}
\end{tikzpicture}
\vspace{-0.5cm}
\caption{Training intent space illustration, with the unseen intent {\tt GetWeather}}
\label{fig:experiment_training}
\end{figure}
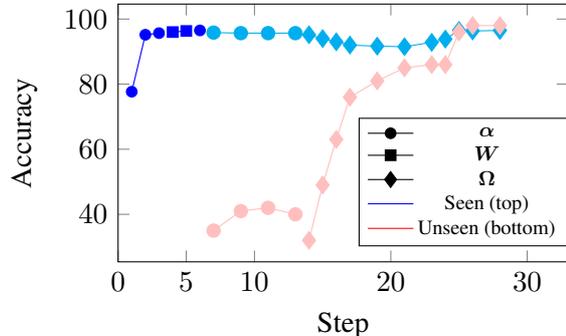

For the first experiment we picked 6 intents as seen, and left {\tt GetWeather} as unseen. The complete list of intents is shown in Table~\ref{tab:unseen_intents}. The model was trained by following the approach described in Figure \ref{fig:intent_space_training}. Figure~\ref{fig:experiment_training} below provides an illustration of the overall training process that includes accuracies achieved while training on seen (step 0-6) and unseen (7-) intents. Note that the second curve that occurs from the step 7 onwards shows accuracies predicting the unseen intent.%
\vspace{-0.3cm}
We start by training first the bases $\bm W$ and then the coordinates $\bm\alpha$. While the majority of the gain comes from training bases this is partially caused by them being trained first, which makes the improvement achieved from training coordinates more limited. By training $\bm W$ first we ensure that each basis learns a good representation. It is however highly likely that by letting bases to learn very fast the coordinates will not be able to deviate much from the initial one-hot encoded representation. Thus, the use of coordinates as a way to learn relationships between training intents may be compromised in such situations.%

Table~\ref{tab:snips_overall} below summarises the overall performance of Euclidean and simplex intent spaces. The baseline RNN shows a similar level of performance. It is interesting to note that although coordinates in the simplex space are constrained, no loss in the generalisation ability is observed compared to the Euclidean space. Given that the former offers more interpretability, all experiments in the following will be based on the simplex space.%
\begin{table}[H]
\centering
\begin{tabular}{|c|cc|}
\hline
\multirow{2}*{Space} & \multicolumn{2}{c|}{Accuracy (\%)}\\
 & Seen & Unseen\\
\hline\hline
$\mathbb E$uclidean & 97.00 & 95.00 \\
$\mathbb S$implex & 97.33 & 95.00 \\
\hline
\end{tabular}
\caption{Performance on the SNIPS dataset}
\label{tab:snips_overall}
\end{table}
\vspace{-0.4cm}
The learnt intent space is illustrated below in Figure~\ref{fig:snips_coordinates}. The strong diagonal shape was expected given large gains in accuracy obtained by training the bases first. Note that even though the intent space is diagonal when an unseen intent is introduced it is highly likely to not attach itself to just one of the training seen intents. Thus, even though the diagonal shape is somewhat concerning it does not necessarily mean that the intent space cannot be used to successfully model unseen intents.%
\begin{figure}[H]
\centering
\tikzset{grid/.style={gray,very thin}}
\begin{tikzpicture}
\begin{axis}[view={0}{90},colorbar, colormap name=whiteblack,height=5cm,width=6cm, , ytick={1, 2, 3, 4, 5, 6}, xtick={1, 2, 3, 4, 5, 6}]
\addplot3[matrix plot] file {data/alpha.dat};
\end{axis}
\end{tikzpicture}
\caption{Intent space coordinates for seen intents}
\label{fig:snips_coordinates}
\end{figure}
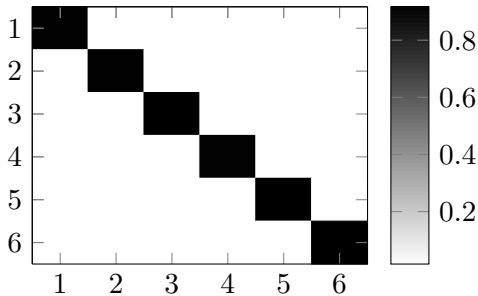
\vspace{-0.3cm}

\subsection{Unseen intents}

\subsubsection{Expansion}
The performance of intent spaces on seen intents is in line with the current state of the art models \citep{zhang-etal-2019-joint}. We would now like to expand it by adding the unseen intent {\tt GetWeather}. We start by adding new coordinates ${\bm\alpha}_{C+1}$ initialised to a uniform distribution over the bases and optimise only these parameters. The bottom curve in Figure~\ref{fig:experiment_training} indicates that optimising only coordinates enables to reach a limit of unseen performance at 40\%. Such level of performance is significantly below exceptions, but above a random classifier.%

Given that the coordinates provide only 6 model parameters the previous result is not surprising. To increase modelling power we expanded the intent space by means of matrix parameters ${\bm\Omega}_{C+1}$ and trained them whilst keeping all other parameters fixed as illustrated by Figure~\ref{fig:intent_space_training}(b). The accuracy improves from 95.67\% on seen and 40.00\% on unseen intents to 97.33\% on seen and 95.00\% on unseen intents. Thus, without any retraining of the previously learnt parameters, the intent space achieves high performance on both the seen and unseen intents. To investigate if the high level of performance on unseen intents is limited to the {\tt GetWeather} intent only, we excluded each of the seen intents in turn. Table~\ref{tab:unseen_intents} shows that the intent space achieves an average performance of 95.36\% on seen and 89.57\% on unseen intents with the weighted average accuracy of 94.53\%.%
\begin{table}[H]
\centering
\begin{tabular}{|c|c|c|c|}
\hline
\multirow{2}*{\#} & \multirow{2}*{Unseen} & \multicolumn{2}{c|}{Accuracy (\%)}\\
 & & Seen & Unseen\\
\hline\hline
1 & {\small\tt AddToPlaylist} & 97.50 & 95.00 \\
2 & {\small\tt BookRestaurant} & 97.17 & 96.00 \\
3 & {\small\tt PlayMusic} & 96.83 & 82.00 \\
4 & {\small\tt RateBook} & 95.00 & 88.00 \\
5 & {\small\tt SearchCreativeWork} & 85.00 & 92.00 \\
6 & {\small\tt SearchScreeningEvent} & 98.67 & 79.00  \\
7 & {\small\tt GetWeather} & 97.33 & 95.00\\
\hline
\end{tabular}
\caption{Accuracy for the SNIPS dataset, with different unseen configurations. The listed intent was unseen, while the remaining ones where seen.}
\label{tab:unseen_intents}
\end{table}
The coordinates of unseen intents learned in each of these configurations are illustrated in Figure~\ref{fig:experiment_unseen_alpha} below, where the $x$ and $y$ axes refer to the numbers shown in the first column of Table \ref{tab:unseen_intents}.%
\begin{figure}[H]
\centering
\begin{tikzpicture}[scale=0.85]
\begin{axis}[view={0}{90},colorbar, colormap name=whiteblack, height=6cm,width=8cm, ytick={1, 2, 3, 4, 5, 6, 7}]
\addplot3[matrix plot] file {data/unseen_alpha.dat};
\end{axis}
\end{tikzpicture}
\caption{Intent space coordinates for unseen intents}
\label{fig:experiment_unseen_alpha}
\end{figure}
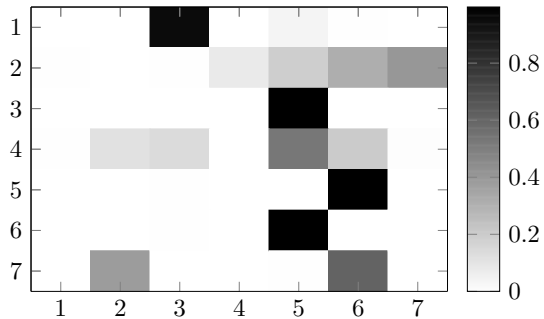
\vspace{-0.3cm}
Figure~\ref{fig:experiment_unseen_alpha} demonstrates that intent space is able to find certain relationship between intents. {\tt AddToPlayList} (1) is being modelled mainly through {\tt PlayMusic} (3), which is exactly what a playlist consist of. {\tt PlayMusic} on the other hand is being modelled by {\tt SearchCreativeWork} (5) which can be explained by the fact that music is a creative art. There are also intents like {\tt BookRestaurant} (2) which consist of a variety of intents, which indicates that it cannot be modelled by a single intent as was expected.%

The previous experiments showed that we are able to get a high performance on unseen intents. We will now look at how much data is required to accomplish that. Table~\ref{tab:my_label} shows the effect that the data has on reaching high prediction accuracy. Even with one training sentence (or point) the intent space achieves better than random performance. At around 500 training sentences the performance reaches 85.00\%. It appears however that sufficiently large quantities of data are needed to predict unseen intents nearly perfect.%
\begin{table}[H]
\centering
\begin{tabular}{|c|c|c|}
\hline
\#Unseen & \multicolumn{2}{c|}{Accuracy (\%)}\\
Sentences & Seen & Unseen\\
\hline\hline
-- & 97.33 & --\\
1 & 91.67 & 60.00\\
10 & 91.50 & 61.00\\
100 & 90.50 & 66.00\\
500 & 88.67 & 85.00\\
1000 & 91.17 & 86.00\\
1500 & 97.33 & 98.00\\
\hline
\end{tabular}
\caption{Performance on the SNIPS dataset with different number of unseen intent sentences} 
\label{tab:my_label}
\end{table}

\subsubsection{Multiple unseen intents}
We have so far seen that it is possible to add one unseen intent.
In the next experiment we will be adding two unseen intents {\tt RateBook} and {\tt BookRestaurant} at the same time whilst keeping the remaining 5 intents as seen. The increase in the complexity caused by adding more than one intent has not led to drop in predicting either seen or unseen intents. The seen performance was 95.60\% while the unseen performance was 97.00\% which is inline with what we were seeing in Table~\ref{tab:unseen_intents} when only one unseen intent was added. Figure~\ref{fig:multiple_intents} below illustrates the coordinates learnt in this experiment and compares them to those trained one at a time.%
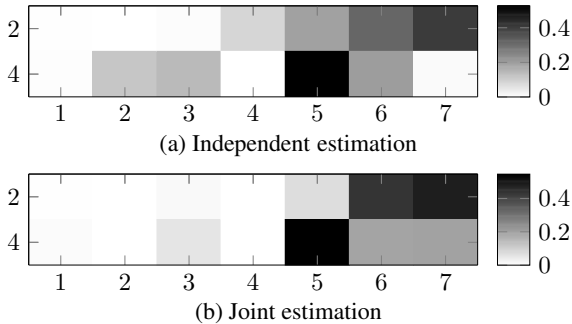
\begin{figure}[H]
\centering
\begin{tikzpicture}[scale=0.85]
\begin{axis}[view={0}{90},colorbar, colormap name=whiteblack, height=3cm, width=8.5cm]
\addplot3[matrix plot] file {data/unseen_alpha_partial.dat};
\end{axis}
\end{tikzpicture}\vspace{-0.2cm}\\
{\small (a) Independent estimation}\vspace{0.2cm}\\
\begin{tikzpicture}[scale=0.85]
\begin{axis}[view={0}{90},colorbar, colormap name=whiteblack, height=3cm, width=8.5cm]
\addplot3[matrix plot] file {data/alpha_2_unseen.dat};
\end{axis}
\end{tikzpicture}\vspace{-0.2cm}\\
{\small (b) Joint estimation}
\caption{Intent space coordinates for {\tt BookRestaurant} (2) and {\tt RateBook} (4)}
\label{fig:multiple_intents}
\end{figure}
\vspace{-0.3cm}
We can see that the main intent contributors are mostly the same, which indicates that the intent space is able to capture the key semantic features of those intents. It also appears that adding one or two intents at the same time makes no significant difference to how the intent space views them.%

\subsubsection{Detection}
Unseen intent detection is an important first step for incorporating unseen intents into intent spaces. Although more advanced approaches are available  \citep{post-processing-unknown-intent, lin-xu-2019-deep, yan-etal-2020-unknown} , the intent spaces can also be examined for unseen intent detection. For simplicity, we focused on the entropy based approach described in equation \eqref{eq:entropy}. The ROC curve that summarises unseen event detection capabilities is shown below in Figure~\ref{fig:roc}.%

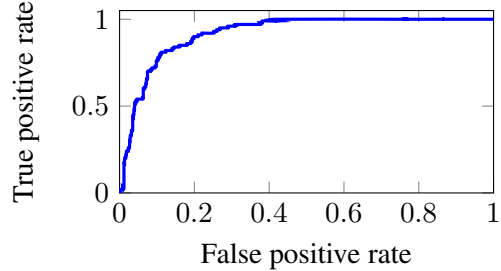
\begin{figure}[H]
\begin{center}
\begin{tikzpicture}
\begin{axis}[xlabel=False positive rate,ylabel=True positive rate,legend style={at={(1.0,0.5)}},xmin=0,xmax=1,ymin=0,ymax=1.05,height=4cm,width=6.5cm]
\addplot[smooth,very thick,blue] file {data/roc_unseen.dat};
\end{axis}
\end{tikzpicture}
\end{center}
\vspace{-0.4cm}
\caption{ROC curve for unseen intent detection}
\label{fig:roc}
\end{figure}
\vspace{-0.3cm}
The ROC curve above shows that the model is able to detect unseen intents better than a random classifier but is far from being perfect. As was discussed in Section~\ref{ssec:unseen_intents} there are more advanced ways to utilise intent space's capabilities.%

\subsubsection{Other datasets}
We also explored the highly popular ATIS dataset to assess intent spaces. The maximum accuracy was 92.91\% for RNN and 93.24\% for intent space, which demonstrates that intent spaces can perform well in situations where the majority of intents have very limited training data. Unfortunately, this task is heavily misbalanced and many sentences of rare intent classes are mislabelled. Therefore, no additional investigation was performed.%

\section{Conclusion}
We are living in a world where new intents constantly appear. Some of them may have a temporary nature while others are here to stay. This provides a strong motivation for investigating approaches for detecting and modelling unseen intents. In this paper we propose a novel approach for representing intents in a continuous space, which is called intent space. The intent space is highly flexible and allows to add intents unseen during training without retraining any of its parameters. This continuous representation allows to investigate relationships between the intents, which is not normally possible with standard approaches. We demonstrate high prediction performance on two separate datasets. The future work will examine the use of more powerful bases representations and alternatives to weighted combination approaches to obtain coordinates.%

\bibliographystyle{acl_natbib}
\bibliography{anthology,acl2021}
\cleardoublepage

\pagenumbering{gobble}
\renewcommand\thesection{\Alph{section}}
\appendix

\section{Datasets}
The table below contains the distribution of intents in the SNIPS dataset used in the experiments. The mapping from integer to intent can be found in table \ref{tab:unseen_intents}.
\begin{table}[H]
\center
\begin{tabular}{|c|c|c|c|}
    \hline
    Intents & Train & Validation & Test \\ 
    \hline
    1 & 1842 & 100 & 100 \\
    \hline
    2 & 1873 & 100 & 100 \\
    \hline
    3 & 1900 & 100 & 100 \\
    \hline
    4 & 1856 & 100 & 100 \\
    \hline
    5 & 1854 & 100 & 100 \\
    \hline
    6 & 1859 & 100 & 100 \\
    \hline
    7 & 1900 & 100 & 100 \\
    \hline\hline
    \textbf{Total Samples} & \textbf{13084} & \textbf{700} & \textbf{700} \\
    \hline
\end{tabular}
\caption{SNIPS intent distribution}
\end{table}

The table below contains the distribution of intents in the ATIS dataset used in the experiments.    
\begin{table}[H]
    \center
    \begin{tabular}{|c|c|c|c|}
    \hline
    Intents & Train & Validation & Test \\ 
    \hline
    Flight & 3666 & 0 & 632 \\
    \hline
    Airfare & 423 & 0 & 48 \\
    \hline
    ground\_service & 255 & 0 & 36 \\
    \hline
    Airline & 157 & 0 & 38 \\
    \hline
    Abbreviation & 147 & 0 & 33 \\
    \hline
    Aircraft & 81 & 0 & 9 \\
    \hline
    Flight\_time & 54 & 0 & 1 \\
    \hline
    Quantity & 51 & 0 & 3 \\ 
    \hline
    Flight+airfare & 21 & 0 & 12 \\
    \hline
    Airport & 20 & 0 & 18 \\
    \hline
    Distance & 20 & 0 & 10 \\
    \hline
    City & 19 & 0 & 6 \\
    \hline 
    Ground\_fare & 18 & 0 & 7 \\
    \hline
    Capacity & 16 & 0 & 21 \\
    \hline
    Flight\_no & 12 & 0 & 8 \\ 
    \hline
    Meal & 6 & 0 & 6  \\
    \hline\hline
    \textbf{Total Samples} & \textbf{4966} & \textbf{0} & \textbf{888} \\
    \hline
\end{tabular}
\caption{ATIS intent distribution}
\end{table}

\section{Experimental validation results}
In the experiments that examined adding two unseen intents validation set accuracy was 95.20\% on seen and 97.00\% for unseen. The tables below replicate the tables given in the main body of the paper with test set accuracies replaced with validation set accuracies.

\begin{table}[H]
\centering
\begin{tabular}{|c|cc|}
\hline
\multirow{2}*{Space} & \multicolumn{2}{c|}{Accuracy (\%)}\\
 & Seen & Unseen\\
\hline\hline
$\mathbb E$uclidean & 96.50 & 98.00 \\
$\mathbb S$implex & 96.33 & 100.00 \\
\hline
\end{tabular}
\caption{Performance on the SNIPS dataset}
\label{tab:snips_overall_validation}
\end{table}

\begin{table}[H]
\centering
\begin{tabular}{|c|c|c|c|}
\hline
\multirow{2}*{\#} & \multirow{2}*{Unseen} & \multicolumn{2}{c|}{Accuracy (\%)}\\
 & & Seen & Unseen\\
\hline\hline
1 & {\small\tt AddToPlaylist} & 96.83 & 95.00 \\
2 & {\small\tt BookRestaurant} & 96.33 & 93.00 \\
3 & {\small\tt PlayMusic} & 97.67 & 84.00 \\
4 & {\small\tt RateBook} & 93.50 & 96.00 \\
5 & {\small\tt SearchCreativeWork} & 85.17 & 86.00 \\
6 & {\small\tt SearchScreeningEvent} & 97.67 & 83.00  \\
7 & {\small\tt GetWeather} & 96.50 & 98.00\\
\hline
\end{tabular}
\caption{Accuracy for the SNIPS dataset, with different unseen configurations. The listed intent was unseen, while the remaining ones where seen.}
\label{tab:unseen_intents_validation}
\end{table}

\begin{table}[H]
\centering
\begin{tabular}{|c|c|c|}
\hline
\#Unseen & \multicolumn{2}{c|}{Accuracy (\%)}\\
Sentences & Seen & Unseen\\
\hline\hline
-- & 96.33 & --\\
1 & 92.33 & 44.00\\
10 & 92.33 & 45.00\\
100 & 92.00 & 62.00\\
500 & 90.00 & 84.00\\
1000 & 91.83 & 87.00\\
1500 & 96.17 & 98.00\\
\hline
\end{tabular}
\caption{Performance on the SNIPS dataset with different number of unseen intent sentences} 
\label{tab:my_label_validation}
\end{table}

In ATIS task the highest training set performance was 99.82\% for intent space and 100.00\% for RNN. Due to highly limited number of sentences available to many intents no validation set was used for this task.

\end{document}